# An Efficient Real Time Method of Fingertip Detection

**Jagdish Lal Raheja\*, Karen Das\*\*, Ankit Chaudhary\*\*\***
*\*Digital Systems Group, Central Electronics Engineering Research Institute (CEERI)/ Council of Scientific and Industrial Research (CSIR), Pilani 333031 INDIA (e-mail:jagdish@ceeri.ernet.in)*
*\*\*Tezpur University, Assam, INDIA (e-mail: karenkdas@gmail.com)*
*\*\*\* BITS, Pilani, INDIA (e-mail: ankitc.bitspilani@gmail.com)*

**Abstract**—Fingertips detection has been used in many applications, and it is very popular and commonly used in the area of Human Computer Interaction these days. This paper presents a novel time efficient method that will lead to fingertip detection after cropping the irrelevant parts of input image. Binary silhouette of the input image was generated, using HSV color space based skin filter and hand cropping was performed based on skin pixel histogram of the hand image. The cropped image would be used to determine the fingertips in the image frame.

## 1. INTRODUCTION

Interactive systems based on gesture recognition needs a real time implementation to work with acceptable performance. In the literature many examples can be find out where gesture is used to control the system based on fingertip detection. Our focus is on hand gesture recognition in natural way without using any marker of sensor based gloves. Many researchers have proposed different methods for dynamic hand gesture recognition using fingertip detection, but several limitations can be seen in these approaches. Garg [**Garg, P. et al., 2009**] uses 3D images in his method to recognize the hand gesture, but this process is complex and also not time efficient. Processing time is very critical factor in real time applications as Ozer [**Ozer, I.B.,, 2005**] states "Designing a real-time video analysis is truly a complex task". Yang [Yang] analyses the hand contour to select fingertip candidates and find peaks in their spatial distribution and checks local variance to locate fingertips. These methods are not invariant to the orientation of the hand. There are other methods, which are using directionally Variant templates to detect fingertips [**Kim, J.M. and Lee, W.K., 2008**], [**Sanghi, et al., 2008**]. Few other methods are dependent on specialized instruments and setup like the use of infrared camera [**Oka, K. et al., 2002**], stereo camera [Ying], a fixed background [**Crowley, J.L., et al., 1995**], [**Quek, F.K.H. et al., 1995**] or use of markers on hand. This paper describes a novel method of motion patterns recognition generated by the hand without any kind of sensor or marker.

The detection of moving fingertips in video needs a fast and robust implementation of method. Many fingertip detection methods are based on hand Segmentation technique because it decreases pixel area which is going to process, by selecting only areas of interest. However most hand segmentation methods cannot do a clearly hand segmentation under some conditions like fast hand motion, cluttered background, poor light condition [**Christian**]. Poor hand segmentation method performance usually invalidates fingertip detection methods. Researchers [**Oka, et al., May 2002**], [**Oka, et al., Dec 2002**], [**Sato, 2000**] uses infrared camera to get a reliable segmentation. Few researchers [**Crowle, et al., 1995**], [**Quek, et al., 1995**], [**Christian**], [**Tomita, et al., 1994**], [**Keaton, et al., 2002**], [**Wu et al., 2000**] limits the degree of the background clutter, finger motion speed or light conditions to get a reliable segmentation in their work. Some of fingertip detection methods cannot localize accurately multidirectional fingertips. Researchers [**Crowley, et al., 1995**], [**Quek, et al., 1995**], [**Brown, et al., 2000**], [**Tomita, et al., 1995**] assumes that the hand is always pointing upward to get precise localization.

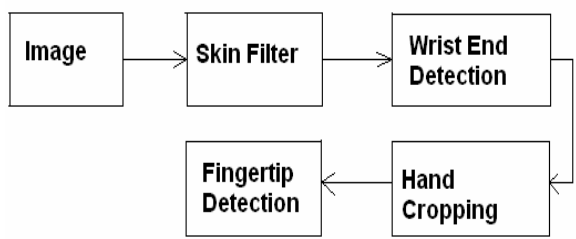

Fig. 1: Algorithm Flow of Fingertip Detection Method

## 2. APPROACH TO FINGERTIP DETECTION

Algorithm flow of Fingertip detection method has been shown in figure 1. It includes five steps. First of all a camera capture a real time video of moving hand in front of system and hand segmentation is done based on skin filter in second step. In the next step wrist end is detected, based on histogram of skin pixels and after this performs hand cropping using different parameters in current image frame of video. Finally fingertips will be detected in the cropped hand image, which is a continuous process for different image frames in the video.

## 2.1 Skin Filter

The skin filter is used on the current input image frame of video. It is based on HSV (can also be based on $YC_bC_r$) colour space. In the HSV colour space the skin would be filtered using the chromacity (hue and saturation) values while in the $YC_bC_r$ colour space, the $C_b$, $C_r$ values would be used for filtering skin. The skin filters are used to create a binary image with background in black colour and the hand region in white. In the next step the binary image need to be smoothened using the averaging filter. Figure 2 shows different steps of skin filtering process. There can be many errors in the output image of skin filter step because of wrong pixel detection or some skin pixels in the background of hand.

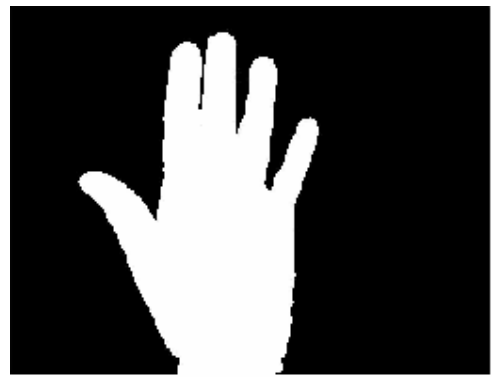

Fig. 3(a): Biggest BLOB

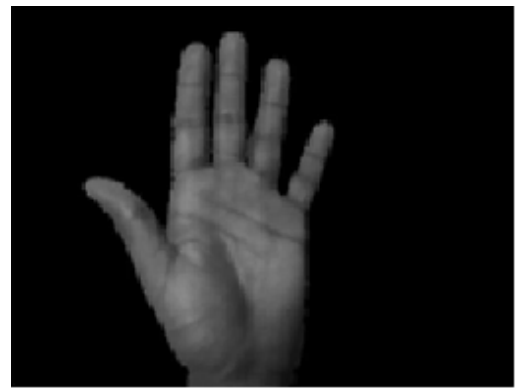

Fig. 3(b): Hand after Filtration

To remove these errors, the biggest BLOB (Binary Linked Object) is considered as the hand and rest the background as shown in figure 3(a). The biggest BLOB represents hand coordinates in '1' and '0' to the background. The filtered out hand is shown in figure 3(b) after removing all errors. The only limitation of this filter is that the BLOB for hand should be the biggest one.

## 2.2 Wrist End Detection

Wrist end detection is based on the histogram of the binary silhouette. Histograms generating functions are:

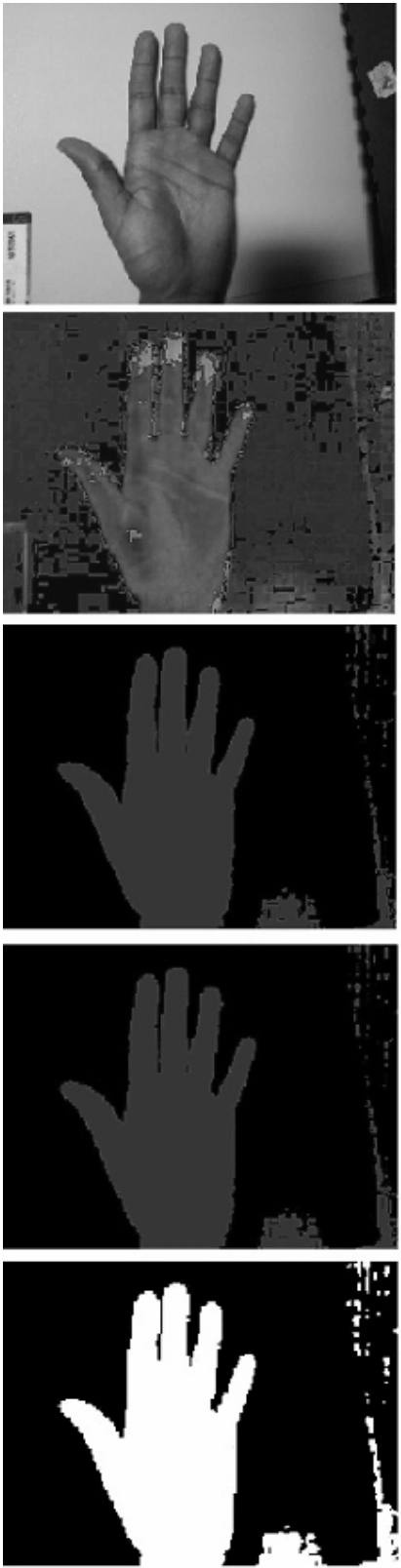

Fig. 2: Skin Filtering Process. Images Shown are (a) Initial Hand Image, (b) HSV Conversion, (c) Filtered Image in HSV format, (d) Smoothen Image after Applying Averaging Filter, (e) Binary Silhouette Respectively.

$$H_x = \sum_{y=1}^{n} imb(x, y)$$

$$H_y = \sum_{x=1}^{m} imb(x, y)$$

Here *imb* represents the binary silhouette and m, n represents the row and columns of the matrix *imb*.

After a 4-way scan of image, we choose the maximum value of 'on' pixels coming out of all scanned ('1' in the binary silhouette). It was noted that maximum value of 'on' pixels represents the wrist end and opposite end of this scan would represent the finger end. Figure 4 shows the scanning process. The yellow bar showed in figure 4 corresponds to the first 'on' pixel in the binary silhouette scanned from the left to right direction. Similarly the green bar corresponds to right to left, red bar corresponds to down to up, and pink bar corresponds to up to downward scan 'on' pixels in the binary silhouette. Now, it is clear that red bar had greater magnitude than other bars for that particular image frame. So we can infer that the wrist end is in downward direction of the frame and consequently the direction of finger is in the upward direction. Here the direction from wrist to finger is known.

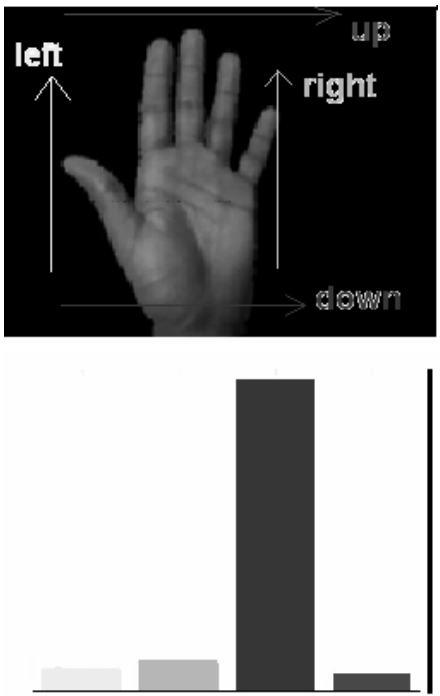

Fig. 4: Image Scanning and Corresponding Bars

### 2.3 Hand Cropping

Hand cropping minimizes the number of pixels to be taken into account for processing which leads to minimization of computation time. In the next step Histogram would be generated from the binary silhouette of the image, as shown in figure 5. It was observed from the histogram that at the point where the wrist ends, a steeping inclination of the magnitude of the histogram starts, whose slope, m can be defined as:

$$m = \frac{y2 - y1}{x2 - x1}$$

As starting point of image where inclination is found, and then the points correspond to the first 'on' pixel scanning from other three sides are found, which gives the coordinates where the image should be cropped. The equations for cropping the image are:

$$imcrop = \begin{cases} origin_{image}, & for\ Xmin < X < Xmax \\ & Ymin < Y < Ymax \\ 0, & elsewhere \end{cases}$$

Where imcrop represents the cropped image, Xmin, Ymin, Xmax, Ymax represent the boundary of the hand in the image.

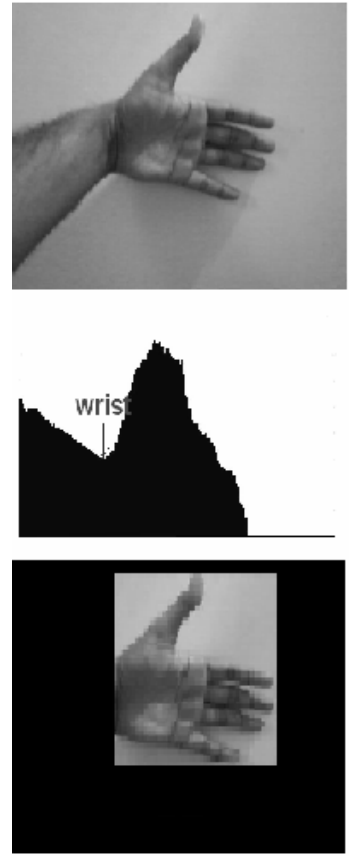

Fig. 5: Hand Cropping. Images shown are (a) Initial Image, (b) Histogram of Binary Silhouette where wrist end can be seen clearly, (c) Cropped Hand Image respectively.

Some results with processing steps for hand cropping are shown in figure 6. The arrows showed in the main frames indicate the directions of scan which were found from wrist end detection step. In all the histograms in figure 6 it is clearly seen that at the wrist point, a steeping inclination starts in the scanning direction.

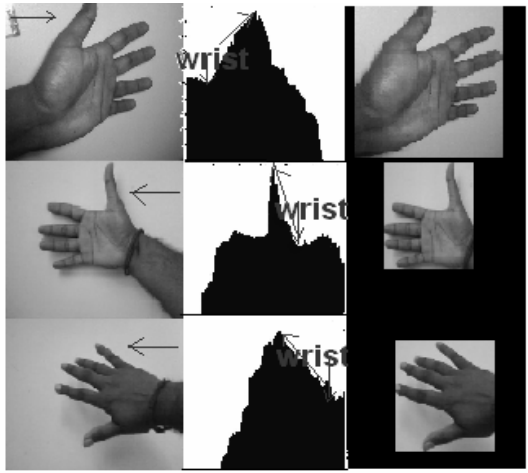

Fig. 6: Results of Hand Cropping Process from Initial Images

*2.4 Fingertip Detection*

Now in the cropped hand image, fingertips will be figured out. Again scanning the cropped binary image and calculate the number of pixels for each row or column based on the hand direction in up-down or left-right. Then intensity values for each pixel are assigned from 1 to 255 in increasing order from wrist to finger end by proportionality. So, each 'on' pixel on the edges of the fingers would be assigned a high intensity value of 255. Now detection of the edge of the fingers is done by just detecting pixels having, intensity of 255. This can be represented mathematically as:

$$pixel_{count}(y) = \sum_{X=Xmin}^{Xmax} imb(x,y)$$

$$modified_{image}(x,y) = round(x * 255 / pixel_{count}(y))$$

$$Finger_{edge}(x,y) = \begin{cases} 1 & if\ modified_{image}(x,y) = 255 \\ 0 & otherwise \end{cases}$$

Here $Finger_{edge}$ gives the boundary of the finger. The line having high intensity pixel, is first indexed and check whether differentiated value lie inside a threshold, if it is then it represents a fingertip. The threshold value changes toward the direction of hand. That threshold can be set after the detection of the direction of hand to the finger which we already know.

### 3. CONCLUSION

The detection of fingertip using a time efficient method has been discussed which will be used in our project 'Controlling the robot using hand gesture'. In this project user will pass controlling information to robot using hand gestures in natural way. The Movement of user's finger will control the robot hand and its working, by moving hand in front of camera without wearing any gloves or markers.